\documentclass[letterpaper]{article} 
\usepackage{aaai2026}  
\usepackage{times}  
\usepackage{helvet}  
\usepackage{courier}  
\usepackage[hyphens]{url}  
\usepackage{graphicx} 
\usepackage{natbib}  
\usepackage{caption} 
\usepackage{algorithm}
\usepackage{algorithmic}
\usepackage{amsmath}
\usepackage{multirow}
\usepackage{algorithm}
\usepackage[misc]{ifsym}
\usepackage{booktabs}
\usepackage{color}

\usepackage{newfloat}
\usepackage{listings}
\DeclareCaptionStyle{ruled}{labelfont=normalfont,labelsep=colon,strut=off} 
\floatstyle{ruled}
\newfloat{listing}{tb}{lst}{}
\floatname{listing}{Listing}
\title{VitaGlyph: Vitalizing Artistic Typography with \\ Flexible Dual-branch Diffusion Models}
\author {
    Kailai Feng\textsuperscript{\rm 1,2},
    Yabo Zhang\textsuperscript{\rm 1},
    Haodong Yu\textsuperscript{\rm 1},\\
    Zhilong Ji\textsuperscript{\rm 2},
    Jinfeng Bai\textsuperscript{\rm 2},
    Hongzhi Zhang\textsuperscript{\rm 1},
    Wangmeng Zuo\textsuperscript{\rm 1},
}
\affiliations {
    \textsuperscript{\rm 1}Harbin Institute of Technology,
    \textsuperscript{\rm 2}Tomorrow Advancing Life\\
}

\begin{document}
\twocolumn[{
\renewcommand\twocolumn[1][]{#1}
\maketitle
\vspace{-35pt}
\begin{center}
  \includegraphics[width=\linewidth, scale=1]{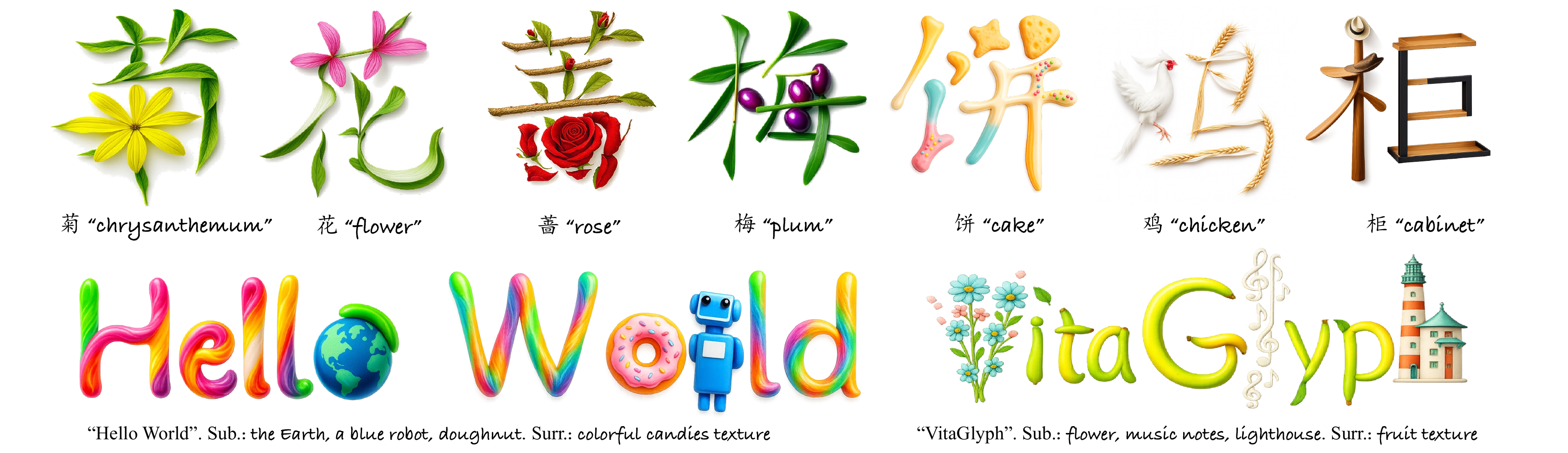}
   \captionof{figure}{\textbf{Examples of our VitaGlyph.} The rendered images not only convey the intrinsic semantics of input characters (first row) but also visualize the user-specific concepts (second row). Sub. denotes the deformable glyph region that conveys the semantics; Surr. denotes the texture in remaining area.
  }
  \label{fig:first}
  \vspace{0pt}
\end{center}
}]

\begin{abstract}
Artistic typography is a technique to visualize the meaning of input character in an imaginable and readable manner. With powerful text-to-image diffusion models, existing methods directly design the overall geometry and texture of input character, making it challenging to ensure both creativity and legibility. In this paper, we introduce a dual-branch, training-free method called \textit{VitaGlyph}, enabling flexible artistic typography with controllable geometry changes while maintaining the readability.
The key insight of VitaGlyph is to treat input character as a scene composed of a \textit{Subject} and its \textit{Surrounding}, which are rendered with varying degrees of geometric transformation.
{To enhance the visual appeal and creativity of the generated artistic typography,} the \textit{subject} flexibly expresses the essential concept of the input character, while the \textit{surrounding} enriches relevant background without altering the shape, {thus maintaining overall readability}.
Specifically, we implement VitaGlyph through a three-phase framework: 
\textbf{(i)} Knowledge Acquisition leverages large language models to design text descriptions for the subject and surrounding.
\textbf{(ii)} Regional Interpretation detects the part that most closely matches the subject description and refines the structure via Semantic Typography.
\textbf{(iii)} Attentional Compositional Generation separately renders the textures of the \textit{Subject} and \textit{Surrounding} regions and blends them in an attention-based manner.
Experimental results demonstrate that VitaGlyph not only achieves better artistry and readability but also manages to depict multiple customized concepts, facilitating more creative and pleasing artistic typography generation. Our code will be made publicly available.
\vspace{-5pt}
\end{abstract}


\section{Introduction}
\label{sec:intro}
\begin{figure*}[t!]
  \centering
  \includegraphics[width=\linewidth]{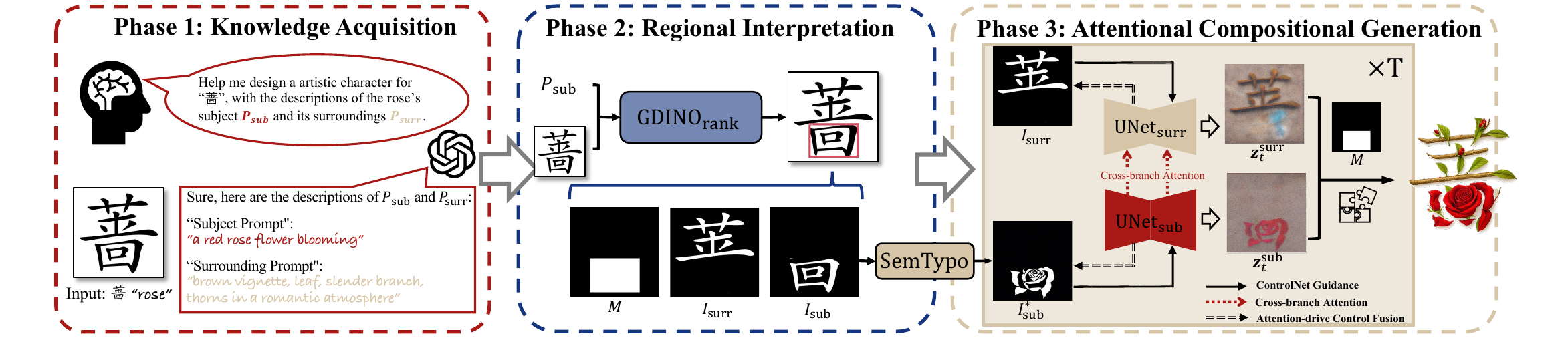}  
  \vspace{-10pt}
  \caption{
    \textbf{Overview of VitaGlyph.} 
    VitaGlyph divides the input glyph image into \textit{Subject} and \textit{Surrounding} and follows a three-phase pipeline: 
    \textbf{(i)} Given an input character, Knowledge Acquisition leverages LLMs to design prompts for both subject and surrounding.
    \textbf{(ii)} With the subject prompt and glyph image, Regional Interpretation detects the part that best matches the subject description and transform the subject's structure via Semantic Typography.
    \textbf{(iii)} Attentional Compositional Generation separately renders textures for both regions through attention-based mechanism.
  }
  \label{fig:pipe}
  \vspace{-15pt}
\end{figure*}

Artistic typography focuses on emphasizing the visual meaning of words while maintaining their legibility. This technique has widespread applications across various fields such as commercial advertising~\cite{gao2023textpainter, lin2023autoposter}, social education~\cite{vungthong2017images, duda2023integrative}, and design promotion~\cite{zhao2023udifftext, xiao2024typedance}. Traditional artistic typography involves manually applying design principles such as color, spacing, and font choice to convey meaning~\cite{he2023wordart,iluz2023word}. However, this process is often time-consuming and lacks the flexibility to dynamically adapt to varying contents and concepts.

Recent works~\cite{iluz2023word,he2023wordart,mu2024fontstudio} leverage the powerful capabilities of pre-trained text-to-image diffusion models~\cite{rombach2022high} to advance the design process. Despite significantly improving productivity and diversity, it remains challenging to flexibly portray the visual characteristics (\emph{e.g.}, the transformation of texture and geometry) of input words while maintaining good readability.
FontStudio~\cite{mu2024fontstudio} focuses solely on depicting the given word through style and texture while neglecting geometric deformation. In contrast, Word-as-image~\cite{iluz2023word} and Khattat~\cite{hussein2024khattat} target black-and-white designs by altering the shape of selected letters. 
WordArt Designer~\cite{he2023wordart} applies distorted transformations to the entire word when attempting to convey multiple concepts, which may sacrifice readability due to significant changes in the overall glyph structure. For example, in Fig.~\ref{fig:qua}, the ``tomato'' resembles several tomatoes more than the character itself. MetaDesigner\cite{he2024metadesigner}, as an integrated system that employs multiple models, requires substantial manual effort and extensive engineering techniques to achieve promising artistic typography.
Moreover, current methods are constrained by their ability to convey only a singular interpretation of a given concept, thereby further restricting flexibility and limiting users' control in the creation of artistic typography.

In this work, we introduce \textit{VitaGlyph} to facilitate flexible and imaginative artistic typography with controllable transformations of glyph geometry while maintaining readability. The core idea of VitaGlyph is to divide an input character (\emph{e.g.}, ``rose'' in Fig.~\ref{fig:pipe}) into the \textit{Subject} (\emph{e.g.}, “red rose”) and the \textit{Surrounding} (\emph{e.g.}, ``brown vignette, leaf, slender branch...''). The \textit{Subject} expresses the essential concept of the input character, while the \textit{Surrounding} enriches the relevant background without altering the shape.

We implement VitaGlyph through a three-phase approach: Knowledge Acquisition, Regional Interpretation, and Typography Stylization. In the first phase, we leverage large language models (\emph{e.g.}, ChatGPT~\cite{chatgpt2023}) to acquire prior knowledge of the input character, designing text prompts for the \textit{Subject} and \textit{Surrounding}. 
In the second phase, different from previous works, we utilize Grounding-DINO~\cite{liu2023grounding} to identify the part of the input glyph that most resembles the subject concept. It allows the decomposed subject inherently contain semantic priors, with which Semantic Typography is used to deform the geometry of the subject image naturally.
During the third phase, we render the subject/surrounding regions with their corresponding prompts through Attentional Compositional Generation(ACG). To better enhance the aesthetic qualities, we inventively propose Cross-branch Attention and Attention-driven Control Fusion.

Extensive experimental results demonstrate that the proposed VitaGlyph offers significant advantages in character transformations while maintaining strong readability. Furthermore, by employing independent diffusion models, VitaGlyph enables artistic text generation with multiple customized concepts—an accomplishment that other approaches fail to achieve. Additionally, the introduction of controllable conditions provides users with greater flexibility in designing personalized content such as signatures and advertisements, as shown in Fig.~\ref{fig:app}.

We summarize our main contributions as follows:
\begin{itemize}
    \item We introduce VitaGlyph, a region‐aware framework for flexible, expressive artistic typography via independent stylization without sacrificing legibility.
    \item We formalize the dual concepts of \textit{Subject} and \textit{surrounding}. \textit{Subject} transforms a glyph’s core structure for artistic expression and \textit{surrounding} only enriches decorative details for preserving readability. 
    \item VitaGlyph integrates Attention‐Guided Compositional Generation, enhancing aesthetic quality and producing more natural blending.
    \item VitaGlyph can be flexibly extended to depict multiple subject concepts, allowing users to add customized descriptions while maintaining intrinsic semantics.
\end{itemize}

\section{Related Work}
\label{sec:rel}
\subsection{Artistic Text Generation}
Artistic text generation~\cite{bai2024intelligent} is divided into text stylization, which focuses on visual effects~\cite{wang2023anything, mao2022intelligent}, and Semantic Typography, which deforms text to match its meaning~\cite{tanveer2023ds, he2023wordart, hussein2024khattat}. Models like eDiff-I\cite{balaji2022ediff}, DeepFloyd~\cite{DeepFloyd_23}, and TextDiffuser-2~\cite{chen2023textdiffuser} enhance generation using LLMs, while others~\cite{xie2023creating,liu2024dynamic,pu2024dynamic} add motion.
However, these methods often capture only one layer of meaning. We propose a dual-branch model that combines stylization and typography for more complex text generation.

\subsection{LLMs for Image Generation}

Large language Models(LLMs)~\cite{achiam2023gpt} have extend their impressive abilities from language tasks to vision synthesis. Most generation-related works~\cite{lian2023llm, yang2024mastering, he2023wordart} use LLMs for prompt understanding, layout planning and typography design. Due to these insights, we incorporate the LLM to design artistic characters.

\subsection{Controllable and Compositional Generation}

ControlNet-based methods~\cite{mou2024t2i, zhang2023adding, huang2023composer, liu2024smartcontrol, zhang2023controlvideo} improve image synthesis control with conditions like segmentation or depth maps, and some studies~\cite{peong2024typographic, yang2024glyphcontrol, mu2024fontstudio} apply these to text generation. Leveraging the geometric nature of glyphs, we use ControlNet with glyph conditions as our foundation.

For multi-concept synthesis, works~\cite{chefer2023attend, agarwal2023star, meral2024conform, rassin2024linguistic} optimize noise maps for concept co-occurrence, while~\cite{jiang2024mc} uses Diffusion Models with LoRAs~\cite{hu2021lora}, and~\cite{liu2023cones} integrates residual embeddings and layout priors.

Few methods~\cite{zhang2024realcompo} combine compositional generation with controllable conditions. Our work achieves pixel-level control in compositional generation using ControlNet and user-specific masks.

\section{Method}

Artistic typography generation task aims to visually convey the meaning of input characters in an appealing and comprehensible way while also ensuring legibility. 
To this end, we introduce a dual-branch method, namely \textit{VitaGlyph}, which performs adaptive rendering on the \textit{Subject} and \textit{Surrounding} components of input glyph.
The \textit{Subject} undergoes structural modifications to flexibly express the intrinsic semantics of the input character, reflecting a strong artistic essence. Meanwhile, the \textit{Surrounding} enriches the details of the subject while maintaining overall readability by adhering to the character's structure throughout the generation process.

As illustrated in Fig.~\ref{fig:pipe}, VitaGlyph consists of three phases: Knowledge Acquisition, Regional Interpretation, and Attentional Compositional Generation.
In the Knowledge Acquisition phase, we ask large language models (\emph{e.g.}, ChatGPT) to acquire the prior knowledge of the \textit{Subject} and \textit{Surrounding} and design their text prompts. 
In Regional Interpretation phase, we parameterize the input character into a glyph image, followed by employing Grounding-DINO (\emph{i.e.}, GDINO) to detect the subject part and Semantic Typography (\emph{i.e.}, SemTypo) to transform the structure.
Finally, during Attentional Compositional Generation~(ACG) phase, we render subject and surrounding images with their corresponding prompts through Cross-branch Attention blending and Attention-driven Control Fusion.

\subsection{Knowledge Acquisition}

The goal of Knowledge Acquisition is to convert input characters into descriptive language prompts for subjects and surroundings that are interpretable by text-to-image diffusion models.
Manually designing these prompts or relying solely on the input characters themselves (\emph{e.g.}, ``rose'') can be challenging, as these methods are either time-consuming or produce descriptions that are difficult to interpret, particularly for abstract words like ``win'' or ``lose''~\cite{he2023wordart}.

large language models such as ChatGPT~\cite{chatgpt2023} have demonstrated significant capabilities in understanding, imagining, and interpreting.
We use ChatGPT~\cite{chatgpt2023} to generate detailed and concrete text descriptions for input characters. However, while related work using LLM focuses primarily on depicting the basic concept of an input character, our approach aims to visualize it more completely and vividly as a compositional scene (\emph{i.e.}, subject and its surroundings).
Specifically, given a query prompt $Q_c$ such as ``Help me design an artistic character for...'', we employ ChatGPT to interpret the input character $c$ into a subject prompt $P_\mathrm{sub}$ and a surrounding prompt $P_\mathrm{surr}$ as follows:
\begin{equation}
    P_\mathrm{sub}, P_\mathrm{surr} = \texttt{GPT}(Q_c, c),
\end{equation}

Supplementary materials provide more detailed examples.

\subsection{Regional Interpretation}
\label{sec:rs}

\begin{figure}[t!]
  \centering
  \includegraphics[width=\columnwidth]{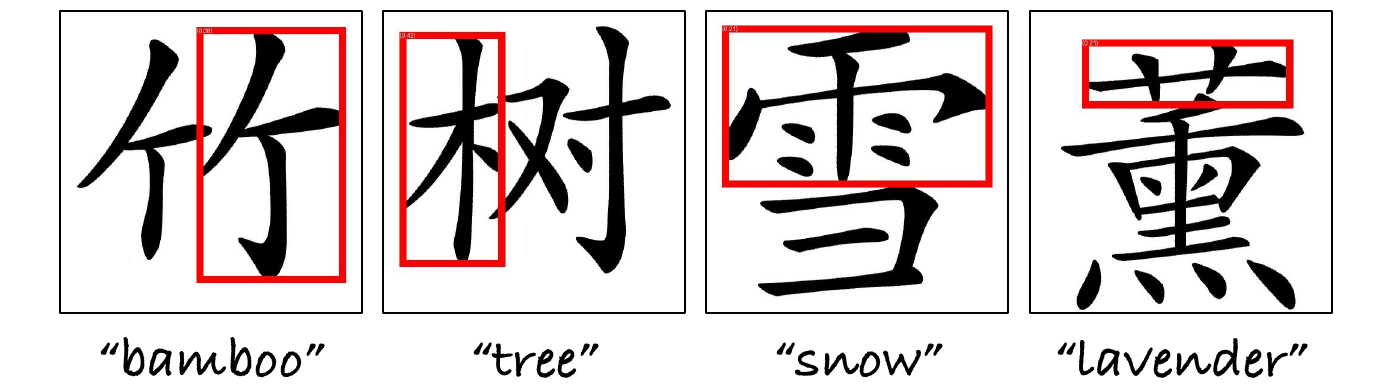}  
  \caption{
    \textbf{Results of Grounding DINO with filtering and ranking strategies.} The red box shows the detected subject region $I_\mathrm{sub}$ that resembles $P_\mathrm{sub}$ most structurally.
    }
  \label{fig:rkgd}
  \vspace{-10pt}
\end{figure}

The Regional Interpretation phase aims to convert the input character $c$ into glyph image $I$ and then divide the image $I$ into subject and surrounding regions at the image level firstly. The subject region corresponds to the part of the image that is most structurally similar to the essential concept described in $P_\mathrm{sub}$. Then it will undergo structural modifications to flexibly express the intrinsic semantics through Semantic Typography (\texttt{SemTypo}).

\subsubsection{Region Selection}
Given an input character $c$, we use ImageFont and ImageDraw of the PIL library to rasterize it into a glyph image $I$ based on a predefined TrueType font.

To preserve the shape of input glyph (\emph{i.e.}, readability) as much as possible, we choose the region most similar to the structure of subject concept as subject region (\emph{e.g.}, ``snow'' in Fig.~\ref{fig:rkgd}).
Compared to previous works that use random region selection~\cite{he2023wordart, he2024metadesigner} or manual region selection~\cite{iluz2023word}, we leverage Grounding-DINO~\cite{liu2023grounding} to detect the part of input image $I$ that matches the subject description $P_\mathrm{sub}$.
Since Grounding-DINO encapsulates rich visual-semantic knowledge, the detected regions inherently contain semantic priors in terms of font structure, thereby efficiently bridging the semantic gap between the selected regions and the subjects, while significantly reducing time. 

As the detected results contain multiple bounding boxes of varying sizes, we filter out those with unreasonable sizes and any that fall below the confidence threshold\footnote{We set the reasonable proportion of the bounding box area relative to the image area between 0.4 and 0.6, and the confidence threshold to 0.5}
Then we select the bounding box with the highest confidence score as the subject region and the remaining of the glyph image as surrounding image:
\begin{equation}
    I_\mathrm{sub}, M = \texttt{GDINO}_\mathrm{rank}(I, P_\mathrm{sub}), I_\mathrm{surr}=I - I_\mathrm{sub},
\end{equation}
where $I_\mathrm{sub}$, $\texttt{GDINO}_\mathrm{rank}$, $M$ and $I_\mathrm{surr}$ denote the subject image, Grounding-DINO with the filtering and ranking strategy, the mask area and the surrounding image, respectively. As shown in Fig.~\ref{fig:rkgd} and Fig.~\ref{fig:GDINO}, due to Grounding-DINO's open-domain detection capabilities, $I_\mathrm{sub}$ (red bounding box) of characters that best resemble $P_\mathrm{sub}$ can be easily detected.

\subsubsection{Semantic Typography}
Previous works~\cite{iluz2023word,he2023wordart} rely on the time-consuming Score Distillation Sampling loss~\cite{poole2022dreamfusion} to transform the structure of input glyphs.
Based on the detected $I_\mathrm{sub}$, we directly use the Depth-to-image model in LDM~\cite{rombach2022high} to adjust the geometry of $I_\mathrm{sub}$, improving efficiency simultaneously.
Specifically, we adopt the SDEdit algorithm~\cite{meng2021sdedit} to perform geometry transformation, conditioned on the depth map $I_\mathrm{sub}$ and the subject prompt $P_\mathrm{sub}$.
It can be described as follows:
\begin{equation}
    I^*_\mathrm{sub} = \texttt{SemTypo}(P_\mathrm{sub}, I_\mathrm{sub}).
\end{equation}

\subsection{Attentional Compositional Generation}
\begin{figure}[t!]
  \centering
  \includegraphics[width=\linewidth]{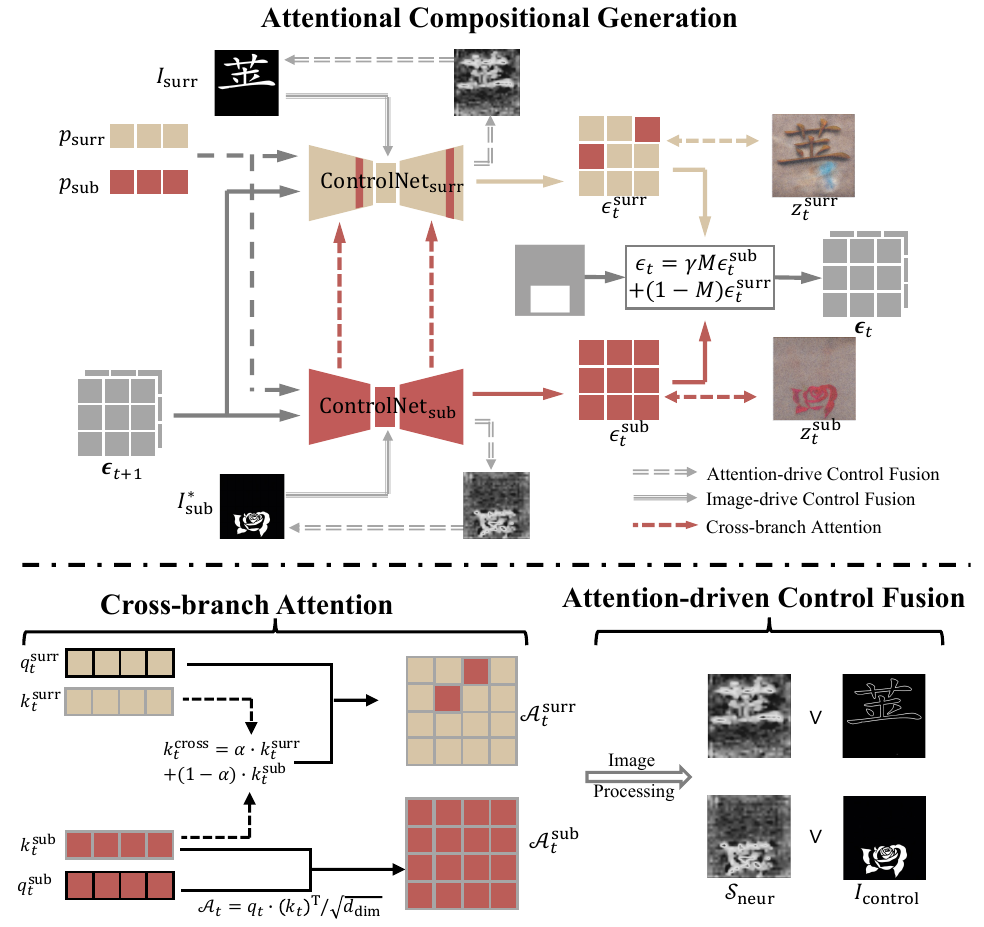}  
  \caption{
    \textbf{Overview of Attentional Compositional Generation.} The ACG module utilizes Stable Diffusion and two independent ControlNets to synthesize and fuse the subject and surrounding regions with the given mask $M$. 
  }
  \label{fig:ACG}
  \vspace{-8pt}
\end{figure}
As mentioned above, the \textit{Subject} and \textit{surrounding} play different roles in depicting the visual semantics of input characters, so their structures and textures are rendered in different ways through two independent ControlNets.
Rather than simple noise‐based blending, we compose them in an attention-based compositional manner termed as Attentional Compositional Generation~(ACG).
It involves Cross-branch Attention mechanism and Attention-driven 
Control Fusion module.

\begin{figure*}[t!]
  \centering
  \includegraphics[width=\linewidth]{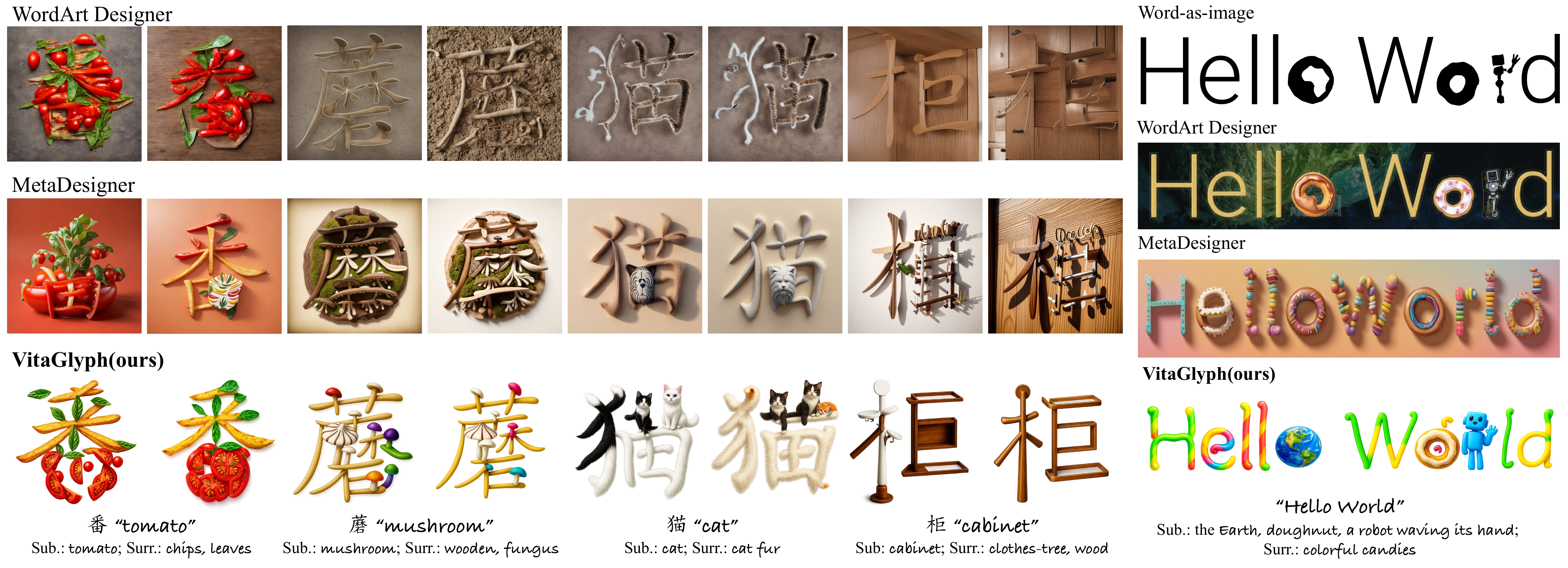}  
  \caption{
    \textbf{Visualization of artistic typography results compared with existing methods.}
    WordArt and MetaDesigner apply transformations to entire characters, often sacrificing readability, as seen in ``mushroom'', where the bottom becomes unrecognizable. 
    Besides, some characters blend into the background without independent diffusion modeling, such as ``cabinet'' in WordArt and ``tomato'' in MetaDesigner. 
    Additionally, former works lack some key semantics (\emph{i.e.}, ''Earth'' in WordArt and ``robot'' in Metadesigner in ``Hello World'').
    In contrast, our method preserves better readability with controllable transformations, separates characters from the background and maintains a visually engaging design. 
  }
  \label{fig:qua}
  \vspace{-10pt}
\end{figure*}

To render subject and surrounding images with their corresponding prompts independently, we apply two ControlNets to separately transfer texture style in a compositional manner.
For the subject image $I_\mathrm{sub}$, after transforming its structure in \texttt{SemTypo} into $I^*_\mathrm{sub}$, we use the segmentation version of ControlNet, termed as $\mathrm{ControlNet}_\mathrm{sub}$. 
Due to the fact that the subject region has semantic priors structurally, the subsequent deformation of this region is more naturally and less likely to suffer significant distortion than that of previous works.
For the surrounding image $I_\mathrm{surr}$, we select the scribble version of ControlNet, denoted as $\mathrm{ControlNet}_\mathrm{surr}$ to maintain the character layout while adding creativity and artistic flair. 
This dual-branch diffusion process decouples the rendering of subject and surrounding, allowing the generated artistic typography to incorporate different concepts.

Specifically, we first encode $P_\mathrm{sub}$ and $P_\mathrm{surr}$ into text embeddings $\boldsymbol{p}_\mathrm{sub}$ and $\boldsymbol{p}_\mathrm{surr}$, respectively.
Both $\mathrm{ControlNet}_\mathrm{sub}$ and $\mathrm{ControlNet}_\mathrm{surr}$ share the same random noise $\epsilon_T \in \mathcal{N}(0,I)$ as the initial latent, conditioned on their corresponding structures and text embeddings.
We use the DDIM algorithm~\cite{song2020denoising} for sampling. To fuse subject and surrounding textures at timestep $t$, given subject mask $M$, we combine the predicted noises $\epsilon^\mathrm{sub}_t$ and $\epsilon^\mathrm{surr}_t$ to compute the overall noise $\epsilon^\mathrm{overall}_t$ as:

\begin{equation}
    \epsilon^\mathrm{overall}_t = \gamma \cdot M \cdot \epsilon^\mathrm{sub}_t + (1-M) \cdot \epsilon^\mathrm{surr}_t,
    \label{eq:fusion}
\end{equation}
where $\gamma \in (0,1)$ is the hyper-parameter to adjust the prominence of the \textit{Subject} within the overall image.

\subsubsection{Cross-branch Attention}
Building on coarse-grained fusion rationale (Eq.~\ref{eq:fusion}), we replace simple noise-based blending~(seen in supplementary materials) with attention-driven fusion. To empower the \textit{Surrounding} branch to carry \textit{Subject} semantics, we insert dedicated cross-branch attention layers into the the $\mathrm{ControlNet}_\mathrm{surr}$ rather than relying on weight mixing. Concretely, during the \textit{Surrounding} denoising process, we extract query-key pairs from $\epsilon_t$ with projection matrices $W_q$ and $W_k$ as
\begin{equation}
    q^\mathrm{surr}_t = \mathcal{F}^\mathrm{surr}_t \cdot W_q, k^\mathrm{surr}_t = \mathcal{F}^\mathrm{surr}_t \cdot W_k,
\end{equation}
where $(q^\mathrm{surr}_t, k^\mathrm{surr}_t)$ represents the query-key pair calculated from intermediate feature map $\mathcal{F}^\mathrm{surr}_t$ at every layer of U-Net. Self-attention scores are generally calculated as $\mathcal{A}^\mathrm{surr}_t = q^\mathrm{surr}_t \cdot (k^\mathrm{surr}_t)^T / \sqrt{d_{dim}}$. Instead, we compose these self-attention scores with cross-attention against \textit{Subject} query-key pairs $(q^\mathrm{sub}_t, k^\mathrm{sub}_t))$. As discussed in \cite{feng2024multi}, self-attention scores naturally encodes significant semantic cues. To leverage this for cross-region interaction, we synthesize the cross-branch key with a hyperparameter $\alpha$\footnote{We set the $\alpha$ as 0.5.} as 
\begin{equation}
    k^\mathrm{cross}_t = \alpha \cdot k^\mathrm{surr}_t + (1-\alpha) \cdot k^\mathrm{sub}_t
\end{equation}
and compute the cross-branch attention:
\begin{equation}
    \mathcal{A}^\mathrm{cross}_t = q^\mathrm{surr}_t \cdot (k^\mathrm{cross}_t)^T / \sqrt{d_{dim}},
\end{equation}
seamlessly replacing the original U-Net attention. This mechanism lets the background location actively query and absorb \textit{Subject} features, yielding a truly semantically fused result that far exceeds vanilla control-map overlays.

\subsubsection{Attention-driven Control Fusion}
We introduce a self-supervised gating scheme that utilizes the UNet's own self-attention at each denoising step to produce a raw attention map, thereby enabling the model to autonomously identify regions that warrant additional emphasis.
This map is then turned into a crisp neural sketch $\mathcal{S}_\mathrm{neur}$ via simple Gaussian smoothing, outlining the key structural skeleton. We fuse this sketch with the original control image in a parameter-free, element-wise manner (\emph{e.g.}, amplifying control features by taking the pixel-wise max), yielding a true attention-driven control signal without any ground-truth supervision. 
\begin{equation}
    I_\mathrm{fused} = [\mathcal{S}_\mathrm{neur} \vee I_\mathrm{control}].
\end{equation}
When combined with our Cross-branch Attention, this attention-driven fusion reinforces \textit{Subject} cues within \textit{Surrounding} regions and refines context coherence, producing sharper strokes and richer textures. And with multiple subjects and concepts, we can extend our framework to multi-concept generation~(seen in supplementary materials).

\section{Experiments}
\begin{figure*}[t!]
  \centering
  \includegraphics[width=\linewidth]{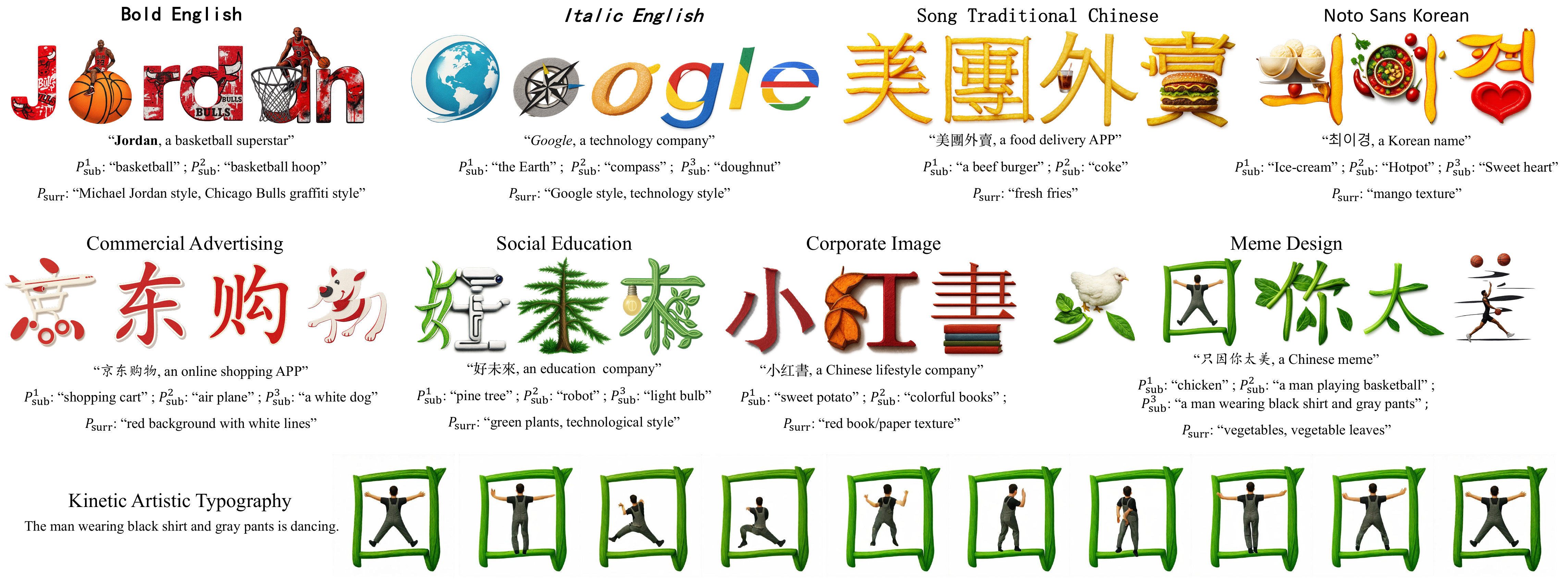}  
  \caption{
    \textbf{Visualization of VitaGlyph's applications.} The experiments show that our method can be applied to other languages and fonts. Additionally, VitaGlyph can be extended to generate with multiple customized concepts and kinetic typography. 
  }
  \label{fig:app}
  \vspace{-5pt}
\end{figure*}

\begin{table*}[t]
    \centering
    \begin{tabular}{l|c|cccc}
    \bottomrule[1.2pt] 
         Methods & Language & CLIP Score $\uparrow$ & GPT-4o(Aes.) $\uparrow$ & GPT-4o(OCR) $\uparrow$ & FID Score $\downarrow$ \\
         \hline
         WordArt Designer~\cite{he2023wordart} & \multirow{2}{*}{Chinese} & 23.93 & 85.12 & 88.33\% & 249.12 \\
         VitaGlyph & & \textbf{26.58} & \textbf{94.35} & \textbf{92.23\%} & \textbf{236.15} \\
         \hline
         Word-as-image~\cite{iluz2023word} & \multirow{2}{*}{English} & {29.01} & {70.66} & {{80.52\%}} & {255.28} \\
         VitaGlyph & & \textbf{{29.16}} & \textbf{{92.18}} & \textbf{{81.96\%}} & \textbf{{247.17}} \\
    \toprule
    \end{tabular}
    \caption{
    \textbf{Quantitative results.} Metrics include CLIP score, GPT-4o Text-Image Matching(TIM), GPT-4o Optical Character Recognition(OCR) and FID Score. Our model outperforms WordArt Designer and Word-as-image by a significant margin.
    }
    \label{tab:main}
    \vspace{-5pt}
\end{table*}

\subsection{Implementation Details}
\subsubsection{Benchmark}
We collect two hundred Chinese words (including single and multiple characters) as an evaluation benchmark and ask ChatGPT to annotate subject and surrounding descriptions. Considering that Word-as-image~\cite{iluz2023word} only generates English words, we also translate the collected benchmark into English. 

More details are provided in the supplementary materials.

\subsubsection{Evaluation Metrics}
We use CLIP~\cite{radford2021learning} image-text similarity to assess similarities between rendered images and prompt descriptions (\emph{i.e.}, $P_\mathrm{sub}$+ $P_\mathrm{surr}$) and FID~\cite{heusel2017gans} scores to assess readability.
Moreover, to better align with human evaluators, we also employ GPT-4o~\cite{gpt4o} to rate the aesthetic scores~(GPT-4o Aes.) and determine whether rendered images and can be correctly recognized as input characters (GPT-4o OCR).

\subsection{Quantitative Comparisons}
As shown in Tab.~\ref{tab:main}, we conduct quantitative experiments on the benchmark mentioned above. For comparisons, we leverage WordArt Designer~\cite{he2023wordart} for Chinese words and Word-as-image~\cite{iluz2023word} for English version. 
Since MetaDesigner has not open-sourced the code and we can only generate results one by one through the website demo. We have only conducted a qualitative comparison with it.

Our VitaGlyph outperforms WordArt Designer in all metrics, especially in OCR with a 3.9\% improvement. We believe this is because, with the controllable and proper transformation of glyph geometry, our method maintains promising legibility. Meanwhile, for English words, VitaGlyph also demonstrates strong generalization capabilities, especially in aesthetic field~(Aes.) with an improvement of 21.52. This further proves that our method can better express essential concepts with elaborately designed subject transformations.

Since all methods' output images are in color while the ground-truth images for FID evaluation are black-and-white, all methods result in high FID scores. However, we still outperform other baselines.

\subsection{Qualitative Comparisons}
We compare VitaGlyph with WordArt Designer and MetaDesigner, both of which are strong baselines for Chinese artistic typography. We also design English words for a qualitative comparison with Word-as-image. 
To ensure a fair comparison, we concatenate $P_\mathrm{sub}$ and $P_\mathrm{surr}$ from our method and use it as input prompt for other works.

In Fig.~\ref{fig:qua}, we see that VitaGlyph renders input characters with higher quality and legibility than WordArt Designer and achieves results that are comparable to, or even better than, those of MetaDesigner. For example, in column $1$, VitaGlyph visualizes the concept of ``tomato'' more concisely, incorporating related elements such as ``fries''. 
However, the deformation regions in WordArt's images are not well defined, leading to unnatural distortions and structural damage. And for those from MetaDesigner, the failure in deformation resulted in the fusion of character and background, ultimately transforming the glyph background into a ``tomato''. 

For the English version, the key concepts for the ``Hello World'' are designed manually as specialties such as ``the Earth, a doughnut, a robot waving its hand and colorful candy texture''. But we can observe the both WordArt and Metadesigner miss the element of ``the Earth''.
As for Word-as-Image, it excels in structural transformation but it can only generate black-white typography. In contrast, VitaGlyph properly generates imaginable artistic typography with the controllable transformation. 
More VitaGlyph examples can be found in supplementary materials. 

\subsection{Ablation Studies}

\begin{table}[t]
    \centering
    \begin{tabular}{l|cccc}
    \bottomrule[1.2pt] 
         {\multirow{2}{*}{Methods}} & {\multirow{2}{*}{CLIP $\uparrow$}} &  \multicolumn{2}{c}{GPT-4o} & {\multirow{2}{*}{FID$ \downarrow$}} \\ \cline{3-4}
         {} & {} & {Aes. $\uparrow$} & {OCR $\uparrow$} & \\ 
         \hline
         w/o $\texttt{GDINO}_\mathrm{rank}$ & 23.90 & 84.47 & 89.15 & 255.34 \\
         w/o \texttt{SemTypo} & 23.53 & 85.79 & \textbf{92.98} & 241.58 \\
         w/o \texttt{Attn} & 25.81 & 92.74 & 92.23 & 239.27 \\
         Ours & \textbf{26.58} & \textbf{94.35} & 92.23 & \textbf{236.15} \\
    \toprule
    \end{tabular}
    \caption{
    \textbf{Ablation of modules in our method.} We provide our ablation studies' quantitative results with different modules removed.
    }
    \label{tab:abl}
    \vspace{-13pt}
\end{table}

\subsubsection{Effect of Region Interpretation}

\begin{figure*}
  \centering
  \includegraphics[width=\linewidth]{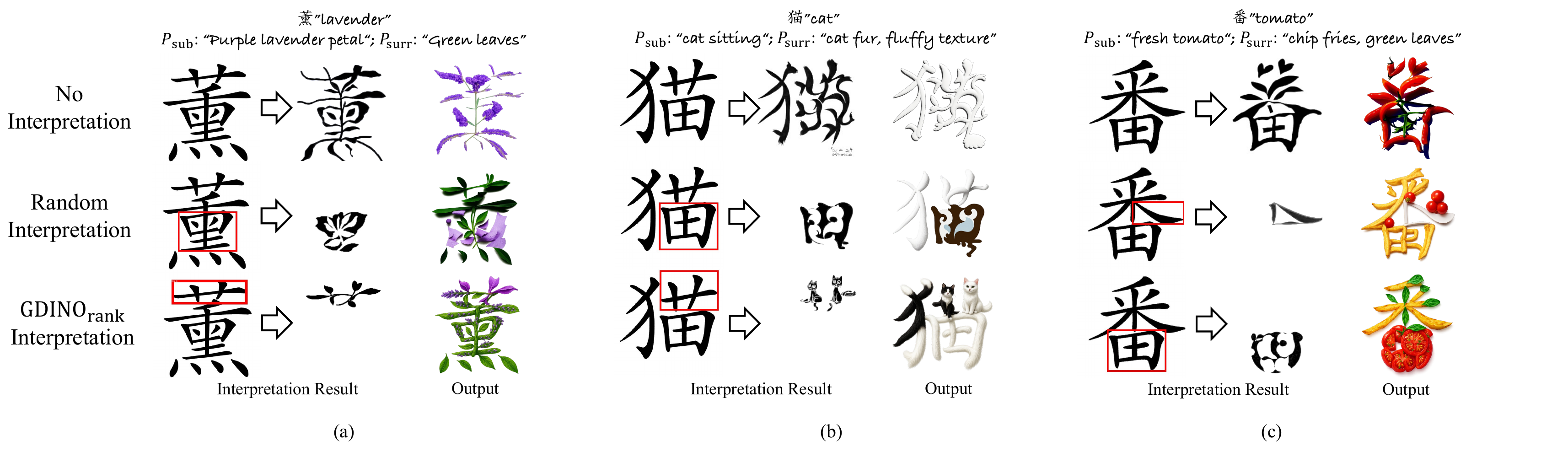}  
  \caption{
    \textbf{Visualization of ablation study on Region Interpretation.}
    \textbf{(a)} Regard the entire image as the subject region.
    \textbf{(b)} Randomly select a part of the image as the subject region (\textit{i.e.}, in \textcolor{red}{red} box). 
    \textbf{(c)} Employ $\texttt{GDINO}_\mathrm{rank}$ to select the most appropriate part as the subject region.
    Results of $I^*_\mathrm{sub}$s and generated images indicate that VitaGlyph with  $\texttt{GDINO}_\mathrm{rank}$ is superior.
  }
  \label{fig:GDINO}   
  \vspace{-5pt}
\end{figure*}

To study the effect of our Region Interpretation from Sec.\ref{sec:rs}, we conduct ablation experiments without $\texttt{GDINO}_\mathrm{rank}$. Specifically, we regard $I$ as $I_\mathrm{sub}$, while keeping the settings for \texttt{SemTypo} unchanged. We solely utilize $\mathrm{controlnet}_\mathrm{sub}$ with $I^*_\mathrm{sub}$ as the additional condition. The input prompt is a concatenation of $P_\mathrm{sub}$ and $P_\mathrm{surr}$. The results are shown in first row of Fig.~\ref{fig:GDINO}. Furthermore, we also randomly select the subject region rather than utilize $\texttt{GDINO}_\mathrm{rank}$, with all other settings unchanged. The results are presented in second row of Fig.~\ref{fig:GDINO}.

Fig.~\ref{fig:GDINO} shows that without Region Interpretation, VitaGlyph faces challenges:
\textbf{(i)}Unreasonable region selection makes it difficult to ensure both prompt alignment and readability simultaneously. For example, the structure of ``lavender'' is deformed (1st row, sub-fig.~a) and the features of ``cat'' are not accurately generated (2nd row, sub-fig.~b).
\textbf{(ii)}The overall deformation of the characters is abnormal and some key concepts, such as ``fries'' in ``tomato'' (1st row, sub-fig.~c), are missing.

For quantitative results, compared to rows 2 and 5 of Tab.~\ref{tab:abl}, the CLIP score improves by 1.91\% and the Aes. score increases by 9.88 wtih 3.08\% in OCR. This is likely because Grounding-DINO identifies the part most resembling the subject prompt, making glyph geometry transformation easier and more natural.

\subsubsection{Effect of Semantic Typography}\label{abl:D2I}

Although Grounding-DINO identifies the part similar to the subject prompt $P_\mathrm{sub}$, Semantic Typography is essential for structural transformation. Ablation experiments without this module show that images fail to capture the character's essence or cause the \textit{Subject} to appear fragmented(see Supplementary Fig.~D). This is due to the gap between $I_\mathrm{sub}$ and real-world objects, which the Depth-to-image module helps bridge.

Based on rows 3 and 5 of Tab.~\ref{tab:abl}, we can conclude that VitaGlyph with Semantic Typography would enrich the semantic expression but sacrifice little readability.

\subsubsection{Effect of Attentional Compositional Generation}
\begin{figure}[t!]
  \centering
  \includegraphics[width=\linewidth]{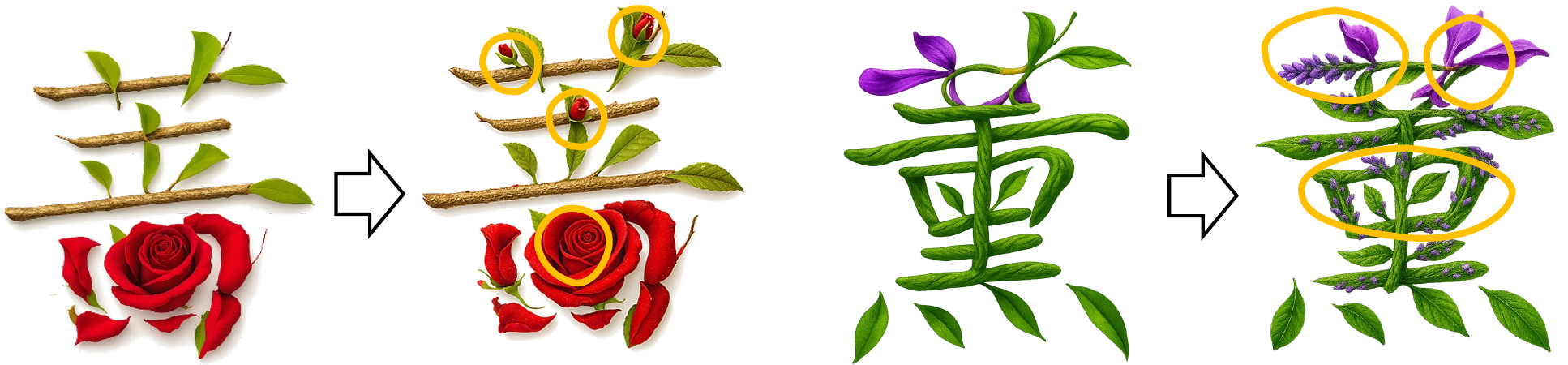}  
  \caption{
    \textbf{Visualization of ablation study on ACG module.} 
  }
  \label{fig:abl_acg}
  \vspace{-8pt}
\end{figure}
To verify the effect of our ACG module, we conduct ablation experiments with noise-based compositional generation. Specifically, we remove the cross-branch attention between \textit{Subject} and \textit{Surrounding} and simplify the attention-driven control as image-driven control, with other settings unchanged. The 4th and 5th row of Tab.~\ref{tab:abl} shows the increase of 1.61 in GPT-4o~(Aes.).
From the Fig.~\ref{fig:abl_acg}, we can also observe more details when ACG module is added.

We validate our ACG module via an ablation that replaces attention-driven fusion with standard noise-based compositional generation (\emph{i.e.}, we disable cross-branch attention and revert from attention-driven back to image-driven control while keeping all other settings fixed.) As reported in Tab.~\ref{tab:abl} (rows 4–5), restoring ACG raises the GPT-4o aesthetic score by 1.61. Qualitative examples in Fig.~\ref{fig:abl_acg} also reveal substantially richer detail when ACG is applied.

\section{Application}

\subsection{Multiple Concepts, Fonts and Languages}
In the first row of Fig.~\ref{fig:app}, we demonstrate VitaGlyph's effectiveness across various fonts and languages, such as ``Bold'' in English, ``Song Traditional'' in Chinese and ``Noto Sans'' in Korean. VitaGlyph not only generates artistic glyph images, but also preserves the original font styles.

Moreover, VitaGlyph is the first work that can extend a single subject to multiple customized subject concepts. As illustrated in Fig.~\ref{fig:app}, it designs subjects like ``the Earth'', ``compass'' and ``colorful doughnut'' within the artistic typography of ``Google''. To address the blurred character details in human generation, we integrate VLM~\cite{gpt4o} to drive fine-grained detail enhancement.
This plays a crucial role in areas such as commercial advertising, brand identity, and corporate image design.

\subsection{Kinetic Artistic Typography}
Our framework adapts to kinetic artistic typography by leveraging I2V models~\cite{wan2025} to animate the \textit{Subject} alone, enabling controllable, semantics-driven glyph motion. Unlike prior methods that emphasize depth or background-to-glyph transitions~\cite{he2023wordart}, we focus on the core motif’s behavior and movement for more expressive animations (Fig. \ref{fig:app}).

\section{Conclusion}
This paper presents VitaGlyph, a novel approach to artistic typography that balances creativity and readability through a controllable glyph transformation. By distinguishing the subject and the surrounding elements, VitaGlyph enhances visual representation while preserving the core meaning of the character. 
Our method makes significant contributions to artistic text generation. 
Future work will focus on further advancements in generative AI to enhance the adaptability of typography across diverse contexts and styles. VitaGlyph will significantly enrich artistic expression, enabling more personalized, dynamic designs.

\clearpage
\twocolumn[
  \begin{center}
    {\LARGE \bfseries Supplementary Materials \par}
    \vspace{1em}
  \end{center}
]
\renewcommand{\thesection}{\Roman{section}}
\setcounter{section}{0}
\renewcommand{\thefigure}{\Alph{figure}}
\setcounter{figure}{0}
\renewcommand{\thetable}{\Alph{table}}
\setcounter{table}{0}

\begin{figure*}[t!]
  \centering
  \includegraphics[width=0.95\linewidth]{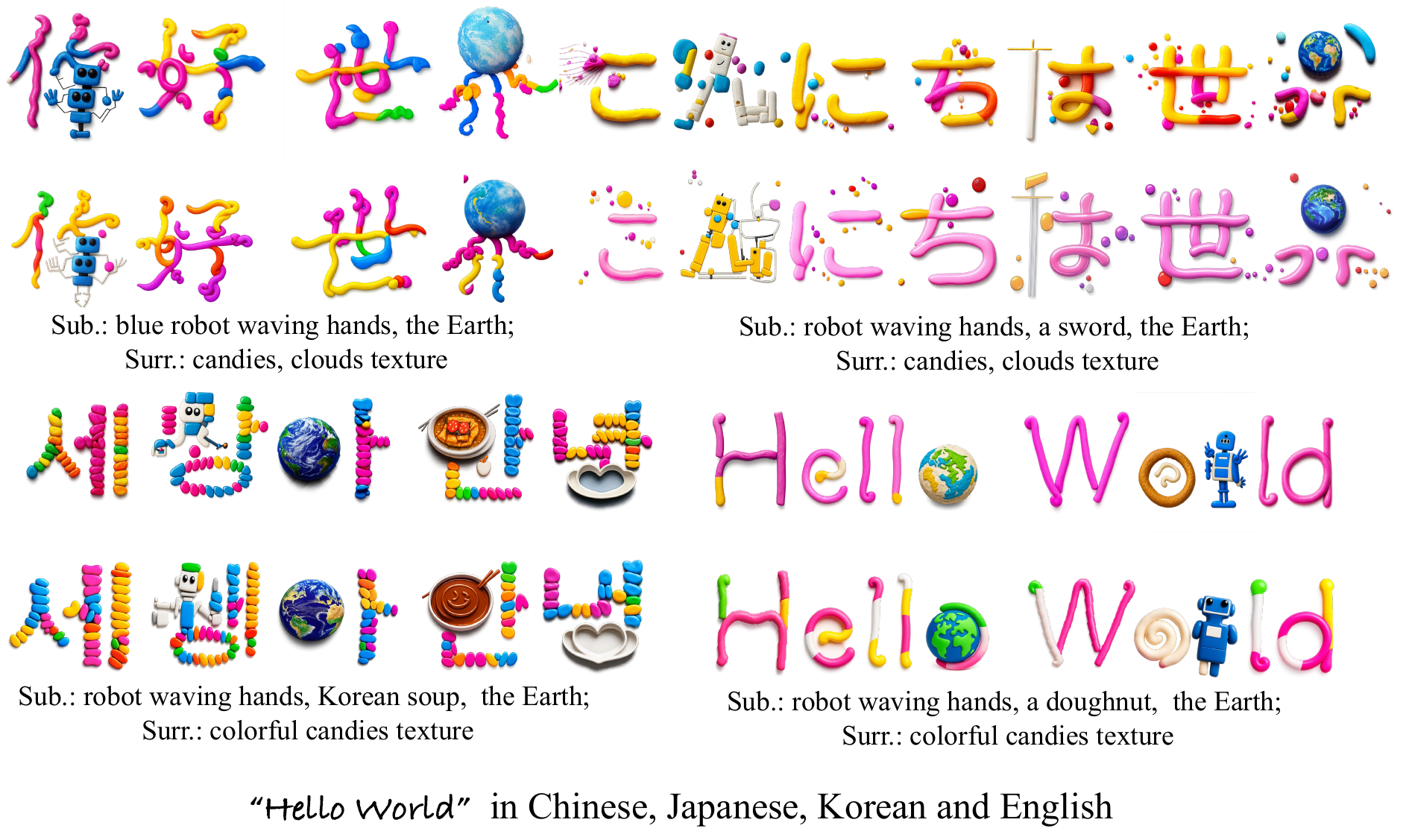}  
  \caption{
    \textbf{More examples of VitaGlyph's application. Our method can be applied to various languages and fonts. With multiple user-specific concepts injected and the utilization of Attentional Compositional Generation, we manage to design slogan like ``Hello World''.}.
  }
  \label{fig:logo}
\end{figure*}

\section{Prompt Examples}
\subsection{LLM Templates:} 
``\textit{Suppose you are a creative and active explainer, dedicated to helping people understand abstract concepts and concrete them. Your task is to identify and elucidate the representative tangible objects within these abstract concepts, helping the audience better connect abstract thinking with real-world entities. 
All questions should follow a standardized format, such as: ``Please list the representative tangible objects in/of \textless CONCEPT \textgreater, along with the appropriate artistic font style.''Your response must adhere to a strict JSON format, specifically: \{``subject prompt'': ``representative tangible object that can express the intrinsic semantic of input \textless CONCEPT \textgreater'', ``surrounding prompt'': ``appropriate artistic font style or texture that can enrich \textless CONCEPT \textgreater.'' \}This format ensures clarity and consistency in responses, making the information easy to parse and understand. 
For example, suppose the question is: ``Please list the representative category or object name in/of `cat', including in real-life, artist works, and film works, and select an appropriate artistic font style.''Your response should be: \{``subject prompt'': ``cake and frosting, sprinkles, layers, with features like sweet, colorful, decadent, multi-tiered.'',``surrounding prompt'': ``texture design is frosting, sprinkles, layers, with creamy, glossy, textured, and delightful details.''\}}''

\subsection{Subject and Surrounding prompt examples}

\noindent \textbf{$Q_{s}$: ``Please list the representative tangible objects in/of plum, along with the appropriate artistic font style.''}

\noindent $A_{s}$: ``\textit{\{``subject prompt'': ``round plum, juicy fruit and vibrant blossom with features such as juicy, round, vibrant, fragrant and delicate.'',
``surrounding prompt'': ``colorful blossom, green leaves, petals, water droplets, velvety skin texture and intricate floral motifs.''\}}''

\noindent \textbf{$Q_{s}$: ``Please list the representative tangible objects in/of rose, along with the appropriate artistic font style.''}

\noindent $A_{s}$: ``\textit{\{``subject prompt'': ``a blooming red rose flower'',
``surrounding prompt'': ``brown vignette, leaf, slender branch, thorns in a romantic atmosphere''\}}''

\section{Noise-based Compositional Generation}
The overall Attentional Compositional Generation process is illustrated in Fig.~4. Since the sampling processes for the \textit{subject} and \textit{surrounding} are conducted independently, and the U-Net outputs are also independent, directly merging them can result in a visually incoherent image, as seen in Fig.\ref{fig:euc}.

In our method, we resolve this fragmentation issue by allowing the subject and surrounding diffusion processes to share the same unconditional noise $\epsilon^\mathrm{uc}_t$ during ACG module.
$\epsilon^\mathrm{overall}_t$ can be further harmonized as follows:
\begin{equation}
    \hat{\epsilon}^\mathrm{overall}_t = \epsilon^\mathrm{uc}_t + s(\epsilon^\mathrm{overall}_\mathrm{t} - \epsilon^\mathrm{uc}_t),
\end{equation}
where $s$ represents the guidance scale and is set to $7.5$ by default.

\section{Multi-concept Generation}
For $N$ subject concepts, we iteratively detect the subject region for the $i$-th concept with \texttt{GroundingDINO}. This can be formalized as follows:
\vspace{-10pt}
\begin{equation}
    I_\mathrm{sub}^i, M^i = \texttt{GDINO}_\mathrm{rank}(I \setminus \bigcup_{j=1}^{i-1} I_\mathrm{sub}^j, P_\mathrm{sub}^i),
\vspace{-10pt}
\end{equation}
Then we obtain multiple subject structures and their corresponding masks, allowing us to achieve a multi-concept scenario. The overall predicted noise is computed as:
\vspace{-10pt}
\begin{equation}
    \epsilon^\mathrm{overall}_t = \sum_{i=1}^N \gamma^i \cdot M^i \cdot \epsilon^\mathrm{sub,i}_t + (1- \sum_{i=1}^N M^i) \cdot \epsilon^\mathrm{surr}_t.
\vspace{-8pt}
\end{equation}

\section{Benchmark Details}
\begin{figure*}[t!]
  \centering
  \includegraphics[width=0.95\linewidth]{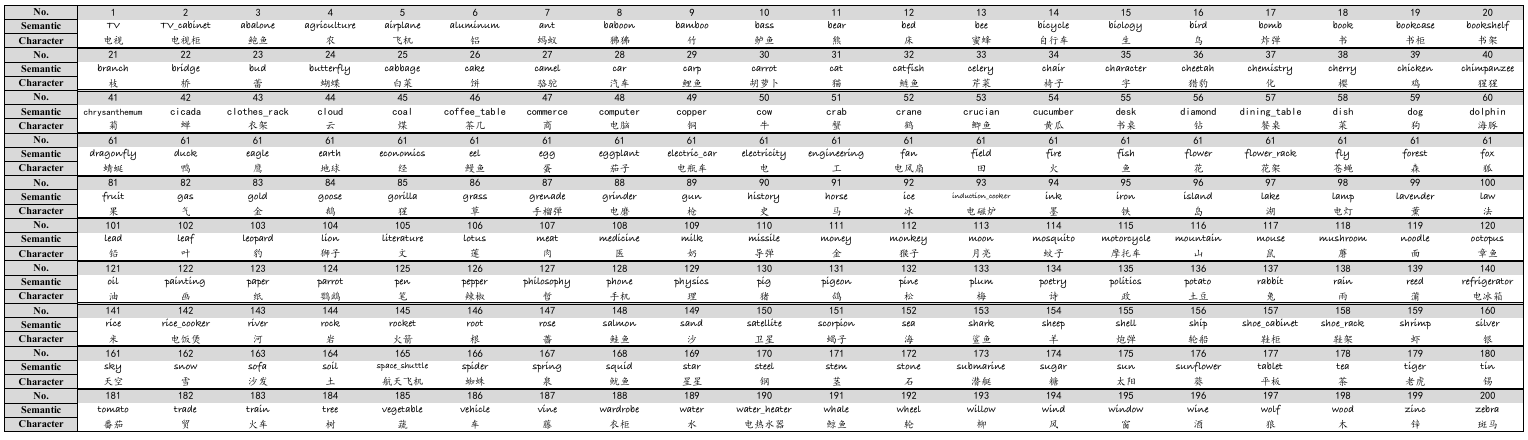}  
  \caption{
    \textbf{Benchmark words of our experiments.}
  }
  \label{fig:samples}
\end{figure*}
We collect 200 Chinese words (including single and multiple characters) as an evaluation benchmark, as shown in Fig~.\ref{fig:samples}.
For every word, we generate 100 samples and we totally create 20000 images as a dataset.
Considering that Word-as-image~\cite{iluz2023word} only generates English words, we also translate the collected benchmark into an English version with 50 samples per word and a total of 10000 images as a dataset.

For words' descriptions, we ask ChatGPT~\cite{chatgpt2023} to annotate subject and surrounding prompts. Since other methods, Word-as-image~\cite{iluz2023word}, WordArt Designer~\cite{he2023wordart} and MetaDesigner~\cite{he2024metadesigner} can only leverage one input prompt, we concatenate the subject and surrounding prompts.

\section{Deployment Details}
In phase Knowledge Acquisition, we use GPT3.5 as our Large Language Model, and the output of the LLM is \textit{JSON} format. 
In phase Regional Decomposition, we use Grounding-DINO with its hyper-parameters all default to the ~\cite{liu2023grounding}. After that, we select the bounding boxes with its confidence over 0.5 and its area proportion of the whole image between 0.4 and 0.6 and then choose the highest confidence as the subject part.
In phase Typography Stylization, we use Stable Diffusion-XL~\cite{podell2023sdxl} as our base model. All generated images are $512\times512$ for both the Depth-to-image and ACG module. 

\section{Effect of LLM’s random outputs}
\begin{table*}[t]
    \centering
    \begin{tabular}{l|c|ccc}
    \bottomrule[1.2pt] 
         Methods & Language & CLIP Score $\uparrow$ & GPT-4o(Aes.) $\uparrow$ & GPT-4o(OCR) $\uparrow$ \\
         \hline
         WordArt Designer~\cite{he2023wordart} & \multirow{2}{*}{Chinese} & 23.93$\pm$1.55 & 85.12$\pm$5.96 & 88.33$\pm$1.95  \\
         VitaGlyph & & 26.58$\pm$1.62 & 94.34$\pm$4.75 & 92.23$\pm$1.93  \\
         \hline
         Word-as-image~\cite{iluz2023word} & \multirow{2}{*}{English} & {29.01$\pm$1.23} & {70.66$\pm$5.78} & {80.52$\pm$1.24} \\
         VitaGlyph & & {29.16$\pm$1.28} & {92.18$\pm$5.26} & {81.96$\pm$1.79}  \\
    \toprule
    \end{tabular}
    \caption{Mean \& Std of quantitative experiments over five turns with different input prompts.
    }
    \label{tab:LLM}
\end{table*}

We perform five regeneration rounds for both $P_\mathrm{sub}$ and $P_\mathrm{surr}$. The results in Tab.~\ref{tab:LLM} show that although ChatGPT’s random output actually impacts the final results, its strong comprehension keeps the deviation within an acceptable range.

\section{More Ablation Studies}
\begin{figure}[t!]
  \centering
  \includegraphics[width=\linewidth]{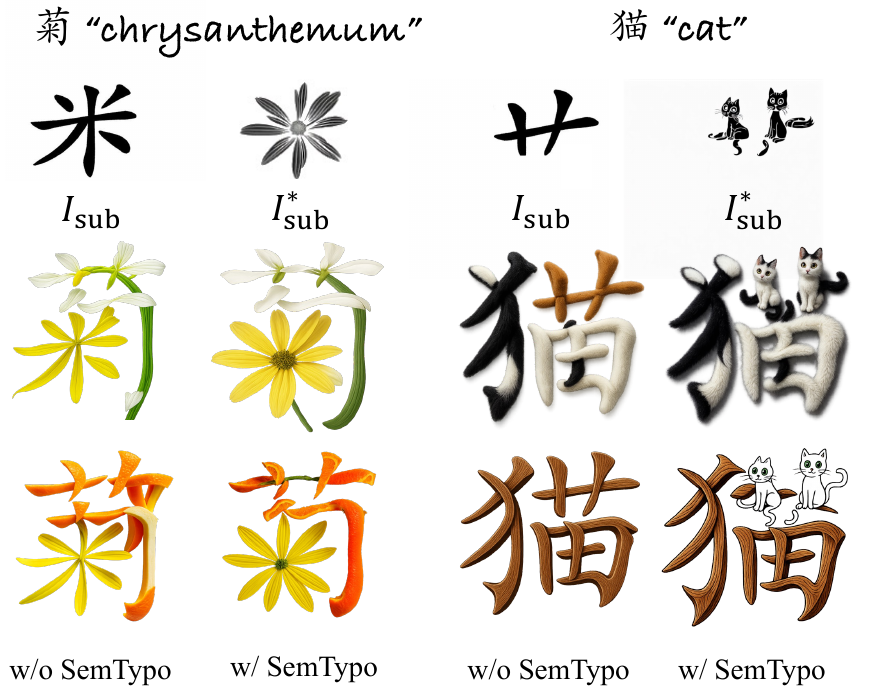}  
  \caption{
    \textbf{Visualization of ablation studies on \texttt{SemTypo}.}}
  \label{fig:d2i}
\end{figure}

\subsection{Effect of shared-$\boldsymbol{\epsilon_T}$}
\begin{figure}[t!]
  \centering
  \includegraphics[width=\linewidth]{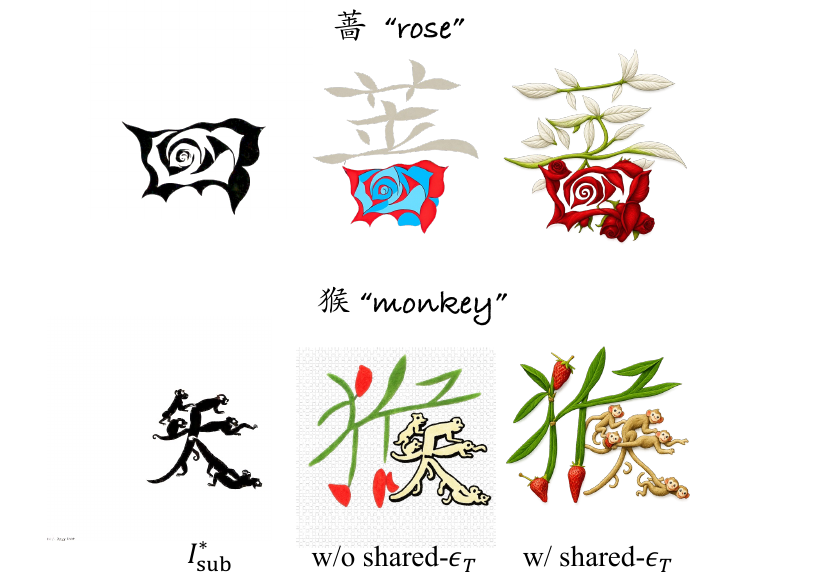}  
  \caption{
    \textbf{Visualization of ablation studies on  shared-$\boldsymbol{\epsilon_T}$.}}
  \label{fig:euc}
\end{figure}

As VitaGlyph employs two ControlNets to individually render subject and surrounding regions, we investigate the impact of shared initial noise (\emph{i.e.}, shared-$\epsilon_T$) on the final results.
Fig.~\ref{fig:euc} indicates that not using shared-$\epsilon_T$ leads to inconsistent appearance and unrealistic lighting.
For example, in the first column of ``rose'' and ``monkey'', the colors of the subject and surrounding regions are visibly incoherent.

\subsection{The Strength of Added Noise in SemTypo}
\begin{figure}[t!]
  \centering
  \includegraphics[width=\linewidth]{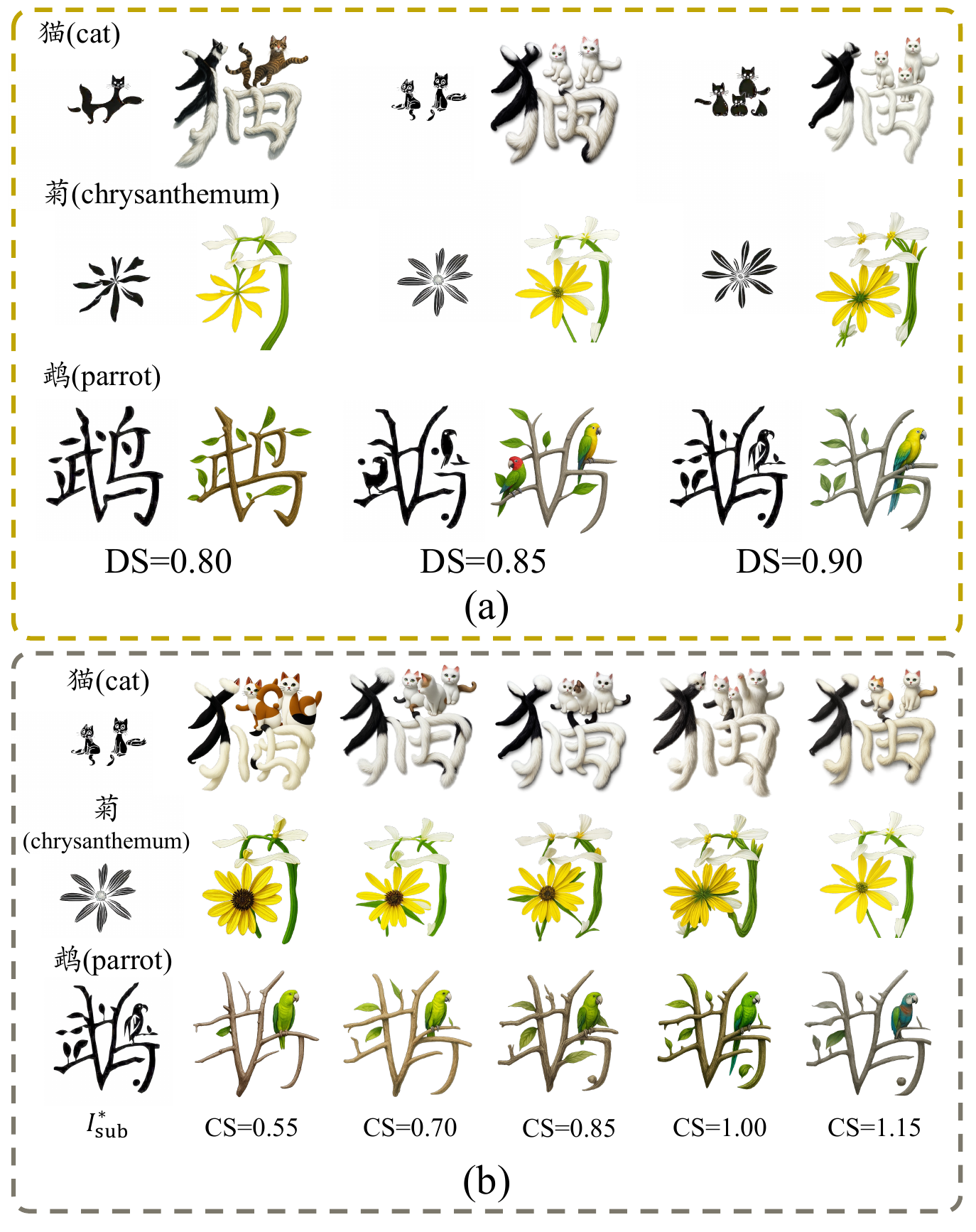}  
  \caption{
    \textbf{Ablation Study on D2I's strength $\mathrm{DS}$ and ControlNet's scale $\mathrm{CS}$.} \textbf{(a)} shows that as $\mathrm{DS}$ increases, the deformation of $I^*_\mathrm{sub}$ grows during \texttt{SemTypo}. \textbf{(b)} demonstrates the generated image morphology aligns more closely with $I^*_\mathrm{sub}$ as $\mathrm{CS}$ increases.
  }
  \label{fig:DSCS}
  \vspace{-10pt}
\end{figure}
Semantic Typography uses the SDEdit algorithm~\cite{meng2021sdedit} based on the Depth-to-image diffusion model to adjust the geometry of the subject region.
As shown in Fig.~\ref{fig:DSCS} (a), as the strength of added noise ($\mathrm{DS}$) increases, more structural details are added to the subject region while it remains in black-and-white color.
When $\mathrm{DS}$ reaches $0.85$, the subject region contains enough details of the given concept, \emph{e.g.}, the ears and tails of cats in the first row.

\subsection{The Conditioning Scale in ControlNet}

We fix $I^*_\mathrm{sub}$ and only adjust the $\mathrm{ControlNet}_\mathrm{sub}$'s guidance scale $\mathrm{CS}$. From Fig.~\ref{fig:DSCS} (b), we observe that as the $\mathrm{CS}$ increases, the generated part of \textit{Subject} becomes more clear and more similar to $I^*_\mathrm{sub}$, shown in row 1. 
However, being too similar to the subject can lead to loss of detail, as seen in row 2 where the flower's stamen disappears from $\mathrm{CS} = 1.0$.

\section{User Study}

\begin{table}[t]
    \centering
    \begin{tabular}{l|ccc}
    \bottomrule[1.2pt] 
         Methods & Text & OCR & Aesthetic \\
         \hline
         WordArt Designer & 29.17\% & 31.67\% & 37.50\% \\
         \hline
         Word-as-image & 25.83\% & 32.50\% & 19.17\% \\
         \hline
         VitaGlyph & \textbf{45.00\%} & \textbf{35.83\%} & \textbf{43.33\%} \\
    \toprule
    \end{tabular}
    \caption{
    \textbf{User Preference Study.} Thirty users participated in the survey, selecting images that best represent the concept (Text Align.), match the character (OCR), and show the highest artistic mastery (Aesthetic).
    }
    \label{tab:user}
\end{table}

We conduct a user study to evaluate our method against WordArt Designer and Word-as-Image on 80 samples, each consisting of an input character and three images from VitaGlyph, WordArt and Word-as-image (random order).
For each sample, we ask 30 workers to select the image that best represents the concept, matches the character, and is the most aesthetically pleasing, yielding 2400 total responses.
The statistical results in Tab.~\ref{tab:user} show that users prefer the rendered images from VitaGlyph across all metrics.

\section{Additional Results}
We provide more examples with multiple customized concepts in Fig.~\ref{fig:logo}. We can observe that we incorporate ``the Earth'', ``doughnut'' and ``robot waving hands'' in artistic typography of ``Hello World'', with its texture style is ``colorful candy texture''. And we can also generate ``Hello World'' in Chinese, Japanese, Korean, and English.



\section{Limitations}
As shown in figure 6, our method would exhibit some deficiencies when generating humans. We believe this is due to the base model, StableDiffusion-1.5. In future work, we will address and refine these issues. Additionally, while we utilize multiple parallel diffusion models for compositional generation, incorporating more than two concepts with different models poses challenges in terms of memory usage and computation time, and we will optimize our implementation as soon as possible.

\section{Future Work}
In addition to addressing the limitations mentioned above, there exist several promising avenues for future research.
\cite{liu2024dynamic, xie2023creating} delve into the animation of artistic words. Offering insights into the harmonious integration of movement and typography artistry, these studies pave the way for us to explore the dynamic and kinetic dimensions of typographic expression. 
Additionally, Our approach is the first to engage artistic typography with compositional generation. We are honored to develop a artistic typography dataset and improve our benchmark with more methods and metrics. This provides a foundation for the development of innovative methods to enhance artistic typography techniques.

\section{Social Impacts}
VitaGlyph has the potential to significantly impact the creative industry by offering new avenues for designers and artists to explore innovative and appealing typographic designs. This technology not only enhances artistic expression and readability, but also has the capacity to enhance user experiences in fields such as advertising, branding, and digital media. Moreover, VitaGlyph's interdisciplinary approach, combining text generation, image processing, and typography design, can foster cross-domain collaborations, driving technological advancements, and fostering innovation. Without proper guidance and ethical considerations, people can engage in harmful practices such as image forgery or digital impersonation. We advocate for the development of legal frameworks that address AI-generated content, including penalties for malicious use.

\section{Time Efficiency}
\begin{table}
    \centering
    \begin{tabular}{l|c}
    \bottomrule[1.2pt]
        Model & Time(Sec.) \\
        \hline
        Word-as-image~\cite{iluz2023word} & 102 \\
        WordArt Designer~\cite{he2023wordart} & 116 \\
        VitaGlyph~(ours) & 15 \\
    \toprule
    \end{tabular}
    \caption{We computed the average time required to generate each image over the course of generating 10 images of the same size on a NVIDIA Tesla V100.}
    \label{tab:time}
    \vspace{-20pt}
\end{table}

Since VitaGlyph operates entirely within the raster domain for both transformation and rendering, it differs from WordArt Designer and Word-as-image, which utilize vector-based methods with SDS Loss~\cite{poole2022dreamfusion} for glyph deformation. This distinction allows us to significantly reduce the inference time, as demonstrated in Tab.~\ref{tab:time}.

\clearpage
\small
\bibliography{aaai2026}
\end{document}